\documentclass[11pt,a4paper]{article}
\usepackage[hyperref]{naaclhlt2019}
\usepackage{times}
\usepackage{latexsym}
\usepackage{dingbat}
\usepackage{url}
\usepackage{graphicx}
\usepackage[font=small,labelfont=small]{caption}

\title{Recommendations for Datasets for Source Code Summarization}
\aclfinalcopy 
\def\aclpaperid{***} 

\author{Alex LeClair, Collin McMillan \\
	Department of Computer Science and Engineering \\
	University of Notre Dame \\
	{\tt \{aleclair, cmc\}@nd.edu} }

\date{}

\begin{document}
\maketitle
\begin{abstract}
  Source Code Summarization is the task of writing short, natural language descriptions of source code.  The main use for these descriptions is in software documentation e.g. the one-sentence Java method descriptions in JavaDocs.  Code summarization is rapidly becoming a popular research problem, but progress is restrained due to a lack of suitable datasets.  In addition, a lack of community standards for creating datasets leads to confusing and unreproducible research results -- we observe swings in performance of more than 33\% due only to changes in dataset design.  In this paper, we make recommendations for these standards from experimental results.  We release a dataset based on prior work of over 2.1m pairs of Java methods and one sentence method descriptions from over 28k Java projects.  We describe the dataset and point out key differences from natural language data, to guide and support future researchers.
\end{abstract}

\parskip=0pt

\section{Introduction}
\label{sec:intro}

Source Code Summarization is the task of writing short, natural language descriptions of source code~\cite{eddy2013evaluating}.  The most common use for these descriptions is in software documentation, such as the summaries of Java methods in JavaDocs~\cite{kramer1999api}.  Automatic generation of code summaries is a rapidly-expanding research area at the crossroads of Computational Linguistics and Software Engineering, as a growing tally of new workshops and NSF-sponsored meetings have recognized~\cite{NL4SEAAAI:2018, nlse2015}.  The reason, in a nutshell, is that the vast majority of code summarization techniques are adaptations of techniques originally designed to solve NLP problems.

A major barrier to ongoing research is a lack of standardized datasets.  In many NLP tasks such as Machine Translation there are large, curated datasets (e.g. Europarl~\cite{europarl}) used by several research groups.  The benefit of these standardized datasets is twofold: First, scientists are able to evaluate new techniques using the same test conditions as older techniques.  And second, the datasets tend to conform to community customs of best practice, which avoids errors during evaluation.  These benefits are generally not yet available to code summarization researchers; while large, public code repositories do exist, most research projects must parse and process these repositories on their own, leading to significant differences on one project to another.  The result is that research progress is slowed as reproducibilty of earlier results is difficult.

Inevitably, differences in dataset creation also occur that can mislead researchers and over or understate the performance of some techniques.  For example, a recent source code summarization paper reports achieving 25 BLEU when generating English descriptions of Java methods with an existing technique~\cite{gu2018deep}, which is 5 points higher than the original paper reports~\cite{iyer2016summarizing}.  The paper also reports 35+ BLEU for a vanilla seq2seq NMT model, which is 16 points higher than what we are able to replicate.  While it is not our intent to single out any one paper, we do wish to call attention to a problem in the research area generally: a lack of standard datasets leads to results that are difficult to interpret and replicate.

In this paper, we propose a set of guidelines for building datasets for source code summarization techniques.  We support our guidelines with related literature or experimentation where strong literary consensus is not available.  We also compute several metrics related to word usage to guide future researchers who use the dataset.  We have made a dataset of over 2.1m Java methods and summaries from over 28k Java projects available via an online appendix (URL in Section~\ref{sec:dataset}).

\section{Related Work}
\label{sec:related}

Related work to this paper consists of approaches for source code summarization.  As with many research areas, data-driven AI-based approaches have superseded heuristic/template-based techniques, though overall the field is quite new.  Work by Haiduc~\emph{et al.}~\cite{haiduc2010supporting, haiduc2010use} in 2010 coined the term ``source code summarization'', and several heuristic/template-based techniques followed including work by Sridhara~\emph{et al.}~\cite{sridhara2010towards, sridhara2011automatically}, McBurney~\emph{et al.}~\cite{mcburney2016automatic}, and Rodeghero~\emph{et al.}~\cite{rodeghero2015eye}.

More recent techniques are data-driven, though the overall size of the field is small.  Literature includes work by Hu~\emph{et al.}~\cite{hu2018deep, hu2018summarizing} and Iyer~\emph{et al.}~\cite{iyer2016summarizing}.  Projects targeting problems similar to code summarization have been published widely, including on commit message generation~\cite{jiang2017automatically, loyola2017neural}, method name generation~\cite{allamanis2016convolutional}, pseudocode generation~\cite{oda2015learning}, and code search~\cite{gu2018deep}.  Nazar~\emph{et al.}~\cite{nazar2016summarizing} provide a survey.

Of note is that no standard datasets for code summarization have yet been published.  Each of the above papers takes an ad hoc approach, in which the authors download large repositories of code and apply their own preprocessing.  There are few standard practices, leading to major differences in the reported results in different papers, as discussed in the previous section. For example, the works by LeClair~\emph{et al.}~\cite{leclair2019icse} and Hu~\emph{et al.}~\cite{hu2018deep} both modify the CODENN model from Iyer~\emph{et al.}~\cite{iyer2016summarizing} to work on Java methods and comments. LeClair~\emph{et al.} and Hu~\emph{et al.} report very disparate results: A BLEU-4 score of 6.3 for CODENN on one dataset, and 25.3 on another, even though both datasets were generated from Java source code repositories.


These disparate results happen for a variety of reasons, such as a difference in data set sizes and tokenization schemes. LeClair~\emph{et al.} use a data set of 2.1 million Java method-comment pairs while Hu~\emph{et al.} use a total of 69,708. Hu~\emph{et al.} also replace out of vocabulary (OOV) tokens in the comments with $<$UNK$>$ in the training, validation, and testing sets, while LeClair~\emph{et al.} remove OOV tokens from the training set only.

\section{Dataset Preparations}
\vspace{0.6cm}
The dataset we use in this paper is based on the dataset provided by LeClair~\emph{et al.}~\cite{leclair2019icse} in a pre-release.  We used this dataset because it is both the largest and most recent in source code summarization.  That dataset has its origins in the Sourcerer project by Lopes~\emph{et al.}~\cite{Lopes+Bajracharya+Ossher+Baldi:2010}, which includes over 51 million Java methods.  LeClair~\emph{et al.} provided the dataset after minimal initial processing that filtered for Java methods with JavaDoc comments in English, and removed methods over 100 words long and comments $>$13 and $<$3 words.  The result is a dataset of 2.1m Java methods and associated comments.  LeClair~\emph{et al.} do additional processing, but do not quantify the effects of their decisions -- this is a problem because other researchers would not know which of the decisions to follow. We explore the following research questions to help provide guidelines and justifications for our design decisions in creating the dataset.

\subsection{Research Questions}

Our research objective and contribution in this paper is to quantify the effect of key dataset processing configurations, with the aim to make recommendations on which configurations should be used.  We ask the following Research Questions:

\vspace{-0.1cm}
\begin{description}
	\item[RQ$_{1}$] What is the effect of splitting by method versus splitting by project?
	
	\vspace{-0.2cm}
	
	\item[RQ$_{2}$] What is the effect of removing automatically generated Java methods?
	
	
\end{description}
\vspace{-0.1cm}

The scope of the dataset in this paper is source code summarization of Java methods -- the dataset contains pairs of Java methods and JavaDoc descriptions of those methods.  However, we believe these RQs will provide guidance for similar datasets e.g. C/C++ functions and descriptions, or other units of granularity e.g. code snippets instead of methods/functions.

The rationale behind RQ$_1$ is that many papers split the dataset into training, validation, and test sets at the unit of granularity under study.  For example, dividing all Java methods in the dataset into 80\% in training, 10\% in validation, and 10\% in testing.  However, this results in a situation where it is possible for code from one project to be in both the testing set and the training set.  It is possible that similar vocabulary and code patterns are used in methods from the same project, and even worse, it is possible that overloaded methods appear in both the training and test sets.  However, this possibility is theoretical and a negative effect has never been shown.  In contrast, we split by project: randomly divide the Java projects into training/validation/test groups, then place all methods from e.g. test projects into the test set.

The rationale behind RQ$_2$ is that automatically generated code is common in many Java projects~\cite{shimonaka2016identifying}, and that it is possible that very similar code is generated for projects in the training set and the testing/validation sets.  Shimonaka~\emph{et al.}~\cite{shimonaka2016identifying} point out that the typical approach for identifying auto-generated code is a simple case-insensitive text search for the phrase ``generated by'' in the comments of the Java files.  LeClair~\emph{et al.}~\cite{leclair2019icse} report that this search turns out to be quite aggressive, catching nearly all auto-generated code in the repository.  However, as with RQ$_1$, the effect of this filter is theoretical and has not been measured in practice.


\begin{table*}[t!]
\centering
    \begin{tabular}{|c|c|c|c|c|c|}
        \hline
        \textbf{SplittingStrategy}&Set1&Set2&Set3&Set4 \\
        \hline
        Split by project&17.81&16.73&17.11&17.99 \\
        \hline
        Split by function&20.97&23.74&23.67&23.68 \\
        \hline
        Auto-generated code included&19.11&19.09&18.04&15.66\\
        \hline
    \end{tabular}
    \caption{Average BLEU Scores from 15 epochs for each of the four sets.}
    \label{table:1}
\end{table*}

\vspace{-0.2cm}
\subsection{Methodology}
\begin{figure}[t!]
	\vspace{-0.5cm}
	\centering
	\includegraphics[width=8cm]{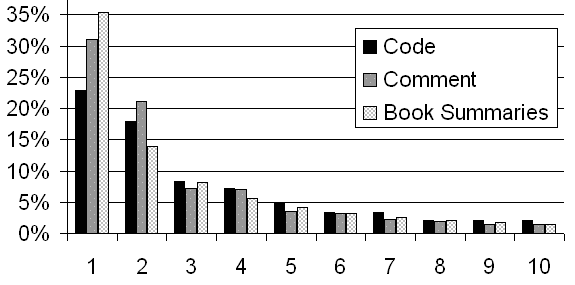}
	\vspace{-0.8cm}
	\caption{Word count histogram for code, comment, and the book summaries.  About 22\% of words occur one time across all Java methods, versus 35\% in the book summaries.}
	\label{fig:rq3wc}
	\vspace{-0.5cm}
\end{figure}
Our methodology for answering RQ$_1$ is to compare the results of a standard NMT algorithm with the dataset split by project, to the results of the same algorithm on the same dataset, except with the dataset split by function.  But because random splits could be ``lucky'', we created four random datasets split by project, and four split by function, seen in Table~\ref{table:2}.  We then use an off-the-shelf, standard NMT technique called {\small\texttt{attendgru}} provided pre-release by LeClair~\emph{et al.}~\cite{leclair2019icse} and used as a baseline approach in their recent paper.  The technique is just an attentional encoder/decoder based on single-layer GRUs, and represents a strong NMT baseline used by many papers.  We train {\small\texttt{attendgru}} with each of the four training sets, find the best-performing model using the validation set associated with that training set (out of 10 maximum epochs), and then obtain test performance for that model.  We report the average of the results over the four random splits.  Note that we used the same configuration for {\small\texttt{attendgru}} as LeClair~\emph{et al.} report, except that we reduced the output vocabulary to 10k to reduce model size.

Our process for RQ$_2$ is similar.  We created four random split-by-project sets in which automatically generated code was \emph{not} removed.  Then we compared them to the four random split-by-project sets we created for RQ$_1$ (in which auto-generated code was removed).

\subsection{Dataset Characteristics}


\begin{figure}[t!]
	\vspace{-0.5cm}
	\centering
	\includegraphics[width=8cm]{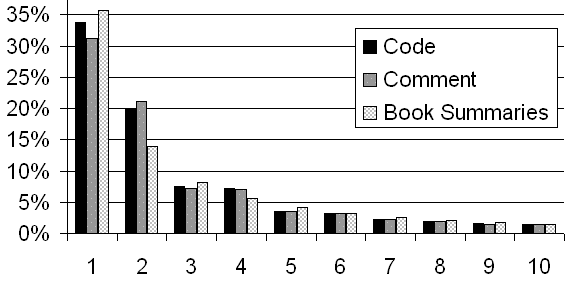}
	\vspace{-0.8cm}
	\caption{Histogram of word occurrences per document.  Approximately 34\% of words occur in only one Java method, 20\% occur in two methods, etc.}
	\label{fig:rq3dc}
	\vspace{-0.5cm}
\end{figure}

We make three observations about the dataset that, in our view, are likely to affect how researchers design source code summarization algorithms.  First, as depicted in Figure~\ref{fig:rq3wc}, words appear to be used more often in code as compared to natural language -- there are fewer words used only one or two times, and in general more used 3+ times.  At the same time (Figure~\ref{fig:rq3dc}), the pattern for word occurrences per document appears similar, implying that even though words in code are repeated, they are repeated often in the same method and not across methods.  Even though this may suggest that the occurrence of unique words in source code is isolated enough to have little affect on BLEU score, we show in Section~\ref{sec:results} that this word overlap causes BLEU score inflation when you split by function. This is important because the typical MT use case assumes that a ``dictionary'' can be created (e.g., via attention) to map words in a source to words in a target language.  An algorithm applied to code summarization needs to tolerate multiple occurrences of the same words. To compare the source code, comments, and natural language datasets we tokenized our data by removing all special characters, lower casing, and for source code -- splitting camel case into separate tokens.

\begin{figure}[b!]
	\vspace{-0.5cm}
	\centering
	\includegraphics[width=8cm]{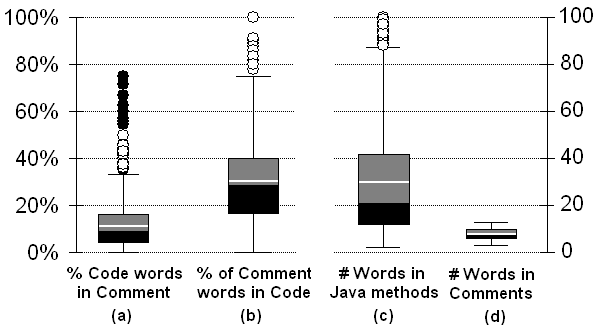}
	\vspace{-0.7cm}
	\caption{Overlap of words between methods and comments (areas a and b).  Over 30\% of words in comments, on average also occur in the method it describes.  About 11\% of words in code, on average, also occur in the comment describing it.  Also, word length of methods and comments (areas c and d).  Methods average around 30 words, versus 10 for comments.}
	\label{fig:rq3box}
	\vspace{-0.6cm}
\end{figure}

A related observation is that Java methods tend to be much longer than comments (Figure~\ref{fig:rq3box} areas (c) and (d)).  Typically, code summarization tools take inspiration from NMT algorithms designed for cases of similar encoder/decoder sequence length.  Many algorithms such as recurrent networks are sensitive to sequence length, and may not be optimal off-the-shelf.

A third observation is that the words in methods and comments tend to overlap, but in fact a vast majority of words are different (70\% of words in code summary comments do not occur in the code method, see Figure~\ref{fig:rq3box} area (b)).  This situation makes the code summarization problem quite difficult because the words in the comments represent high level concepts, while the words in the source code represent low level implementation details -- a situation known as the ``concept assignment problem''~\cite{biggerstaff1993concept}.  A code summarization algorithm cannot only learn a word dictionary as it might in a typical NMT setting, or select summarizing words from the method for a summary as a natural language summarization tool might.  A code summarization algorithm must learn to identify concepts from code details, and assign high level terms to those concepts.

\vspace{-0.2cm}
\section{Experimental Results \& Conclusion}
\label{sec:results}
\vspace{-0.3cm}
In this section, we answer our Research Questions and provide supporting evidence and rational.

\vspace{-0.1cm}
\subsection{RQ$_1$: Splitting Strategy}
\begin{figure}[b!]
	\centering
	\includegraphics[width=8cm]{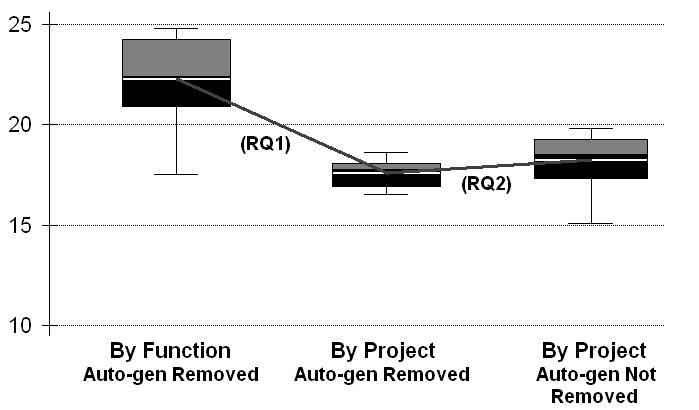}
	\vspace{-0.8cm}
	\caption{{\small Boxplots of BLEU scores from {\small\texttt{attendgru}} for four runs under configurations for RQ$_1$ and RQ$_2$.}}
	
	\label{fig:rq1rq2}
\end{figure}

We observe a large ``false'' boost in BLEU score when split by function instead of split by project (see Figure~\ref{fig:rq1rq2}).  We consider this boost false because it involves placing functions from projects in the test set into the training set -- an unrealistic scenario.  An average of four runs when split by project was 17.41 BLEU, a result relatively consistent across the splits (maximum was 18.28 BLEU, minimum 16.10).  In contrast, when split by function, the average BLEU score was 23.02, and increase of nearly one third as seen in Table~\ref{table:1}.  Our conclusion is that splitting by function is to be avoided during dataset creation for source code summarization.  Beyond this narrow answer to the RQ, in general, any leakage of information from test set projects into the training or validation sets ought to be strongly avoided, even if the unit of granularity is smaller than a whole project.  We reiterate from Section~\ref{sec:intro} that this is not a theoretical problem: many papers published using data-driven techniques for code summarization and other research problems split their data at the level of granularity under study.

\subsection{RQ$_2$: Removing Autogen. Code}

\begin{table*}[t!]
    \centering
    \begin{tabular}{|c|c|c|c|c||c|}
        \hline
        &BP Set1&BP Set2&BP Set3&BP Set4&BF All Sets \\
        \hline
        Training Set&1,935,860&1,950,026&1,942,291&1,933,677&1,943,723 \\
        \hline
        Validation Set&105,693&100,920&104,837&105,997&107,984 \\
        \hline
        Testing Set&107,568&98,175&101,993&109,447&107,984\\
        \hline
    \end{tabular}
    \caption{Number of method-comment pairs in the train, validation, test sets used in each random split set when split by project (BP) and by function (BF).}
    \label{table:2}
    \vspace{-0.4cm}
\end{table*}

We also found a boost in BLEU score when not removing automatically generated code, though the difference was less than observed for RQ$_1$.  The baseline performance increased to 18 BLEU when not removing auto-generated code, and it varied much more depending on the split (some projects have much more auto-generated code than others).  Our recommendation is that, in general, reasonable precautions should be implemented to remove auto-generated code from the dataset because we do find evidence that auto-generated code can affect the results of experiments.

\vspace{-0.1cm}
\section{Discussion}

\vspace{-0.1cm}

This paper provides benefits to researchers in the field of automatic source code summarization in two areas. First, we provide insight into the effects of splitting a Java method and comment dataset by project or by function, and how these different splitting methods effect the task of source code summarization. Second, we provide a dataset of 2.1m pairs of Java methods and one sentence method descriptions in a cleaned and tokenized format (discussed in \ref{sec:dataset}) as well as a training, validation, testing split. 

Note however that there may be cases where researchers wish to adapt our recommendations for a specific context. For example, when generating comments in an IDE. The problem of code summarization in an IDE is slightly different than what we have presented, and would benefit from including code-comment pairs from the same project. IDEs have the advantage of access to a programmer's source code and edit history in real time -- they do not rely on a repository collected post-hoc.  Moreno~\emph{et al.}~\cite{Moreno:2013:Jsummarizer}  take advantage of this information to generate Java class summaries in an eclipse plugin -- their tool uses both the class and project level information from completed projects to generate these summaries, while not using any information from outside sources. 

However, even in this case, care must be taken to avoid unrealistic scenarios, such as ensuring that the training set consists only of code older than the code in the test set. For example, consider a programmer at revision 75 of his or her project who requests automatically generated comments from the IDE, then goes on to write a total of 100 revisions for the project. An experiment simulating this situation should only use revisions 1-74 as training data -- revisions 76+ are ``in the future'' from the perspective of the real world situation. 
\section{Downloadable Dataset}
\label{sec:dataset}

In our online appendix we have made three downloadable sets available as seen in Table \ref{table:3}. The first is our SQL database, generated using the tool from McMillan~\emph{et al.}~\cite{mcmillan2011portfolio}, that contains the file name, method comment, and start/end lines for each method, we call this dataset our ``Raw Dataset''. We also provide a link to the Sourcerer dataset \cite{springerlink:10.1007/s10618-008-0118-x} which is used as a base for the dataset in LeClair~\emph{et al.}~\cite{leclair2019icse}. In addition to the Raw Dataset, we also provide a ``Filtered Dataset'' that consists of a set of 2.1m method comment pairs. In the Filtered Dataset we removed auto-generated source code files, as well all method's that do not have an associated comment. No preprocessing was applied to the source code and comment strings in the Filtered Dataset. The third downloadable set we supply is the ``Tokenized Dataset''. In the Tokenized Dataset, we processed the source code and comments from the Filtered Dataset identically to the tokenization scheme described in Section 5 of~\cite{leclair2019icse}. This set also provides a training, validation, and test set as well as a script to easily reshuffle these sets. 

\vspace{0.2cm}
The URL for download is: 

\textbf{\url{http://leclair.tech/data/funcom}}

\begin{table}[h!]
\centering
\begin{tabular}{|c|c|c|}
        \hline
        Dataset&Methods&Comments \\
        \hline
        Raw Dataset&51,841,717&7,063,331 \\
        \hline
        Filtered Dataset&2,149,121&2,149,121 \\
        \hline
        Tokenized Dataset&2,149,121&2,149,121\\
        \hline
\end{tabular}
\caption{Datasets available for download.}
\vspace{-0.6cm}
\label{table:3}

\end{table}

\vspace{-0.2cm}
\section*{Acknowledgments}

{This  work  is  supported  in  part  by  the  NSF  CCF-1452959, CCF-1717607, and CNS-1510329 grants. Any opinions, findings, and conclusions expressed herein are the authors and do not necessarily reflect those of the sponsors}
\bibliography{main}
\bibliographystyle{acl_natbib}

\end{document}